\documentclass{article}

\usepackage[preprint]{neurips_2024}

\usepackage[utf8]{inputenc}
\usepackage[T1]{fontenc}
\usepackage{hyperref}
\usepackage{url}
\usepackage{booktabs}
\usepackage{amsfonts}
\usepackage{amsmath}
\usepackage{nicefrac}
\usepackage{microtype}
\usepackage{xcolor}
\usepackage{graphicx}
\usepackage{subcaption}
\usepackage{algorithm}
\usepackage{algpseudocode}
\usepackage{float}
\usepackage{tikz}
\usetikzlibrary{positioning}

\makeatletter
\renewcommand{\@noticestring}{Preprint.}
\makeatother

\title{Where Should Optimizer State Live? Tiered State Allocation for Memory-Efficient Mixture-of-Experts Training}

\author{%
  Nuemaan Malik\\
  Independent Researcher\\
  \texttt{nuemaan.research@gmail.com}
}

\begin{document}

\maketitle

\begin{abstract}
Optimizer state is the largest single line item in the memory budget of mixture-of-experts (MoE) training. On a 6.78B-parameter MoE language model AdamW keeps 50.6\,GB of first and second moments to update 12.6\,GB of bfloat16 weights. We study \emph{SkewAdam}, an optimizer built on the observation that the three parameter populations of an MoE differ enough in size and gradient statistics that they should not receive the same state. Those populations are the dense backbone, the experts and the router. SkewAdam keeps float32 momentum plus a factored second moment for the backbone (5\% of parameters), a factored second moment alone for the experts (95\%) and an exact second moment for the router ($<$0.01\%). The resulting state occupies 1.29\,GB or 2.6\% of AdamW's and peak training memory falls from 81.4\,GB to 31.3\,GB, within the budget of a 40\,GB accelerator. In a controlled comparison from identical initializations over 82M tokens, SkewAdam reaches validation perplexity 108.4, ahead of AdamW (126.8), Muon (120.2) and Lion (393.7), and settles router load balance to within 1\% of its uniform floor. The allocation is not what earns that perplexity. A tier ablation reaches the same value while carrying twenty times the state, so the tiers buy memory rather than accuracy. Same-platform runs separate what does earn it. Removing momentum costs 31 perplexity points (tuned Adafactor, 139.7) and replacing the factored second moment and its update clipping with a full second moment costs 10 (tuned AdamW, 118.5), so neither tuned baseline reaches the untuned tiered policy. Where optimizer state lives, these results suggest, matters at least as much as how much of it there is.
\end{abstract}

\section{Introduction}
\label{sec:intro}

Sparse mixture-of-experts (MoE) architectures decouple parameter count from per-token compute \citep{shazeer2017outrageously,fedus2022switch,jiang2024mixtral}. The 6.78B-parameter model we study activates roughly 440M parameters per token. Optimizer state enjoys no such discount. AdamW \citep{kingma2015adam,loshchilov2019decoupled} keeps two float32 moments for every parameter, 8 bytes of state shadowing every 2-byte bfloat16 weight, so our model carries 50.6\,GB of optimizer state on top of 12.6\,GB of weights and training peaks at 81.4\,GB.\footnote{Throughout, GB denotes GiB ($2^{30}$ bytes), matching \texttt{torch.cuda.max\_memory\_allocated}.} The expert bank owns most of those parameters and is also the population whose individual parameters are touched least often.

Memory-efficient optimizers exist, but they treat the network as a homogeneous block. Lion \citep{chen2023symbolic} keeps one momentum buffer and updates with its sign. That halves state but discards gradient magnitude everywhere including in the router, where relative magnitudes carry the load-balancing signal. Muon \citep{jordan2024muon} also keeps a single buffer and orthogonalizes each update by Newton--Schulz iteration, a structural prior designed for dense hidden layers. Adafactor \citep{shazeer2018adafactor} factors the second moment into row and column statistics. That suits a 16.8M-parameter expert matrix well but applies the same recipe to the 0.5M-parameter router that decides where every token goes. None of these methods asks whether different parts of an MoE deserve different state.

SkewAdam does. It allocates optimizer state by tier (Figure~\ref{fig:tiers}). The dense backbone of embeddings, attention and the dense feed-forward block holds 5\% of parameters and sees every token, so float32 momentum is affordable there. It also gets a factored second moment. The experts hold 95\% of parameters, but under top-2 routing over 128 experts each one processes on average $\nicefrac{1}{64}$ of the tokens. They keep only a factored second moment, which removes the single largest state cost in the model. The router keeps an exact, unfactored second moment. Per-logit adaptivity costs 2\,MB and steers all the traffic.

The tiered allocation policy and its closed-form memory model give 1.29\,GB of optimizer state on a 6.78B-parameter MoE and peak training memory of 31.3\,GB. We then run a controlled single-GPU comparison against AdamW, Lion and Muon under identical initialization, identical data order and an identical bfloat16 weight-update path with dithered stochastic rounding for all four. SkewAdam attains the best validation perplexity (108.4) at a throughput within 1.5\% of the fastest baseline. Same-protocol follow-ups on two further GPUs add Adafactor and a GaLore-style baseline, ablate the policy's tiers and sweep the baselines' learning rates. The policy matches full-momentum perplexity at a twentieth of the state and its lead over the baselines survives their tuning. We also analyze routing stability by measuring the load-balancing loss against its analytic floor, finding that Lion's magnitude-blind updates cost perplexity rather than balance, and that Muon drifts away from balance late in training.

\begin{figure}[t]
\centering
\begin{tikzpicture}[
  font=\small,
  tier/.style={draw=black!60, rounded corners=2pt, align=left, inner sep=6pt, text width=3.8cm, minimum height=2.7cm}
]
\node[tier, fill=blue!7] (bb) at (0,0) {%
  \textbf{Backbone} \hfill 341M (5.0\%)\\[1pt]
  {\footnotesize embeddings, attention, dense FFN}\\[5pt]
  $m$: \textbf{fp32} --- 1.27 GB\\
  $V$: \textbf{factored} --- 0.7 MB};
\node[tier, fill=orange!10, right=5mm of bb] (ex) {%
  \textbf{Experts} \hfill 6{,}442M (95.0\%)\\[1pt]
  {\footnotesize 128 SwiGLU experts}\\[5pt]
  $m$: \textbf{none}\\
  $V$: \textbf{factored} --- 12.6 MB};
\node[tier, fill=green!8, right=5mm of ex] (rt) {%
  \textbf{Router} \hfill 0.5M (0.008\%)\\[1pt]
  {\footnotesize top-2 gate, fp32}\\[5pt]
  $m$: \textbf{none}\\
  $V$: \textbf{full fp32} --- 2.1 MB};
\node[below=2.5mm of ex.south, align=center, font=\small] {%
  Total optimizer state: \textbf{1.29 GB} \;---\; 2.6\% of AdamW's 50.55 GB on the same model.};
\end{tikzpicture}
\caption{Tiered state allocation on our 6.78B-parameter MoE ($m$: momentum buffer, $V$: second-moment estimate). Momentum is kept where gradients are dense, factored variance where parameters are plentiful but sparsely excited, and the 2\,MB gate keeps an exact second moment.}
\label{fig:tiers}
\end{figure}

\section{Related work}
\label{sec:related}

\paragraph{Memory-efficient optimizers.}
Adafactor \citep{shazeer2018adafactor} is the closest relative of this work. The factored estimator inside SkewAdam is Adafactor's nonnegative rank-one estimator written with row and column means instead of sums, and the update-RMS clipping is Adafactor's as well. We claim no novelty for either. What we take up is the allocation question that uniform recipes leave open. Adafactor either drops momentum everywhere or keeps it everywhere and factors every matrix regardless of its role. Adam-mini \citep{zhang2025adammini} is the closest precedent for structure-aware allocation: guided by the block Hessian structure of Transformers, it averages Adam's second moment within parameter blocks, head-wise for queries and keys and row-wise elsewhere, halving Adam's footprint while keeping momentum everywhere. Our tiers vary a different axis, which statistics each MoE population keeps at all rather than the granularity of one of them. The two are complementary in that Adam-mini's Hessian principle could set the within-tier granularity. SM3 \citep{anil2019memory} and 8-bit optimizers \citep{dettmers2022bit} compress state by other means and ZeRO \citep{rajbhandari2020zero} shards it across devices rather than shrinking it. All are compatible with a tiered policy and orthogonal to it. GaLore \citep{zhao2024galore} projects gradients into a low-rank subspace before applying Adam, again uniformly across matrices.

\paragraph{Sign- and geometry-based updates.}
Lion \citep{chen2023symbolic} reduces state to a single buffer by updating with the sign of an interpolated momentum. Muon \citep{jordan2024muon} replaces adaptive scaling with Newton--Schulz orthogonalization of the momentum and has been scaled to large dense and MoE models \citep{liu2025muon}. Its authors route embeddings and other non-matrix parameters to an Adam-style rule, which our implementation follows. Both methods are uniform over hidden matrices, expert or not.

\paragraph{MoE training.}
Sparsely gated MoE layers were introduced by \citet{shazeer2017outrageously} and scaled by GShard \citep{lepikhin2021gshard} and Switch Transformers \citep{fedus2022switch}, whose load-balancing auxiliary loss we use. The router z-loss follows ST-MoE \citep{zoph2022stmoe}, which documents how fragile router training can be. Mixtral \citep{jiang2024mixtral} and DeepSeekMoE \citep{dai2024deepseekmoe} are recent open systems. This literature concentrates on architecture and losses. Optimizer state allocation for MoE has received little direct attention.

\paragraph{Low-precision training.}
Mixed-precision practice keeps float32 master weights \citep{micikevicius2018mixed}. Pure-bfloat16 training is attractive for memory but loses small updates to rounding \citep{kalamkar2019study}. Stochastic rounding repairs much of the damage \citep{gupta2015deep,zamirai2021revisiting}. We train with bfloat16 master weights and a dithered approximation to stochastic rounding (Section~\ref{sec:method}), applied identically under every optimizer we compare.

\section{SkewAdam: tiered state allocation}
\label{sec:method}

\paragraph{Factored second moments.}
For a weight matrix $W \in \mathbb{R}^{n \times m}$ with gradient $G_t$, SkewAdam maintains exponential moving averages of the row-wise ($\mathrm{mean_c}$, over columns) and column-wise ($\mathrm{mean_r}$, over rows) mean squared gradient,
\begin{equation}
R_t = \beta_2 R_{t-1} + (1-\beta_2)\,\mathrm{mean_c}(G_t \odot G_t) \in \mathbb{R}^{n}, \quad
C_t = \beta_2 C_{t-1} + (1-\beta_2)\,\mathrm{mean_r}(G_t \odot G_t) \in \mathbb{R}^{m},
\label{eq:factored}
\end{equation}
and reconstructs the second-moment matrix as the rank-one estimate $\widehat{V}_t = R_t C_t^{\top} / \bar{R}_t$, where $\bar{R}_t$ is the mean of $R_t$. This is exact whenever the true second-moment matrix has rank one and coincides with Adafactor's estimator \citep{shazeer2018adafactor}. Storage falls from $nm$ to $n+m$ floats per matrix, so a $4096 \times 4096$ expert matrix needs 32\,KB rather than 64\,MB.\footnote{Stacked expert tensors of shape $(E, n, m)$ factor along the last two axes. In our model the experts are separate matrices, so the two-dimensional path is the one exercised.}

\paragraph{The tiers.}
The policy assigns state by parameter role (Figure~\ref{fig:tiers}, Algorithm~\ref{alg:skewadam}). \emph{Backbone} tensors (embeddings, attention projections, the dense feed-forward block and norms) carry float32 momentum and a factored second moment. Their gradients are dense because every token contributes every step, so momentum smooths a signal that is actually there. At 5\% of parameters the buffer costs 1.27\,GB. \emph{Expert} tensors carry only the factored second moment. They are 95\% of parameters, each expert sees roughly $\nicefrac{1}{64}$ of tokens under top-2-of-128 routing and a momentum buffer here would cost 24\,GB to smooth gradients that arrive sparsely and with high variance. Adafactor's results suggest momentum can be dropped without losing adaptivity \citep{shazeer2018adafactor}. \emph{Router} weights keep a full, unfactored second moment and no weight decay. Factoring a $128 \times 4096$ gate would pool scale statistics across the very logits whose relative magnitudes determine load balance. Exactness here costs 2\,MB. The policy is a judgment about where each ingredient of Adam earns its memory rather than a theorem, and Section~\ref{sec:results} tests it.

\paragraph{Update clipping and low-precision updates.}
The preconditioned update is clipped to unit root-mean-square, $U \leftarrow U / \max(1, \mathrm{RMS}(U))$, Adafactor's update clipping with threshold $1$. Master weights are bfloat16. Each step is computed in float32 and written back through a dithered rounding. Uniform noise of one-ULP width, $\xi \sim \mathcal{U}(-\mathrm{ulp}(w)/2,\, \mathrm{ulp}(w)/2)$ with $\mathrm{ulp}(w) \approx 2^{-7}|w|$, is added before the cast, approximating unbiased stochastic rounding \citep{gupta2015deep,zamirai2021revisiting}. Every optimizer in our comparison uses this same write-back path, so none is privileged by precision handling.\footnote{A side effect we accept deliberately: with $\eta = 3 \times 10^{-4}$ and $\lambda = 0.05$, the decoupled weight-decay step changes weights by $1.5 \times 10^{-5}$ relative, more than two orders of magnitude below the bfloat16 ULP of $2^{-7}$, so weight decay rounds to a no-op in \emph{all} runs. The comparison is fair but effectively unregularized. See Section~\ref{sec:limitations}.}

\paragraph{Memory model.}
Let $N_{\mathrm{bb}}$, $N_{\mathrm{ex}}$, $N_{\mathrm{rt}}$ be the tier sizes. SkewAdam's state is
\begin{equation}
S = 4\big[ N_{\mathrm{bb}} + \textstyle\sum_{W \in \mathrm{bb} \cup \mathrm{ex}} (n_W + m_W) + N_{\mathrm{rt}} \big] \,\text{bytes} \approx 4\,N_{\mathrm{bb}},
\end{equation}
since the factored vectors are negligible. With $N_{\mathrm{bb}} = 341.4$M, $N_{\mathrm{ex}} = 6{,}442.5$M, $N_{\mathrm{rt}} = 0.52$M this gives 1.29\,GB (a component-level account is in Appendix~\ref{app:memory}), against $8 N_{\mathrm{total}} = 50.55$\,GB for AdamW. The saving is structural. It does not depend on quantization and would compound with it.\footnote{Quantized state also carries a scale ceiling that factoring avoids. Current 8-bit optimizer kernels terminate the process, a hard exit from C++ rather than an exception, once a single parameter tensor reaches $2^{31}$ elements, a size that fused expert weights reach in larger MoE models (bitsandbytes issue 1785). We measured the boundary on an A100. The 8-bit step succeeds at $2^{31}-1$ elements and kills the process at $2^{31}$, while factored state built from native PyTorch ops with 64-bit indexing crosses it cleanly. Scripts and logs are in the repository's \texttt{experiments/} directory.}

\begin{algorithm}[t]
\caption{SkewAdam step for one tensor $W$ in tier $\tau$ ($\beta_1{=}0.9$, $\beta_2{=}0.999$, $\epsilon{=}10^{-8}$)}
\label{alg:skewadam}
\begin{algorithmic}[1]
\State $G \gets \nabla_W \mathcal{L}$ in float32
\If{$\tau = \textsc{router}$ \textbf{or} $W$ is a vector}
  \State $V \gets \beta_2 V + (1-\beta_2)\, G \odot G$; \quad $D \gets \sqrt{\max(V, \epsilon)}$ \Comment{full second moment}
\Else
  \State update $R, C$ by Eq.~\eqref{eq:factored}; \quad $D \gets \sqrt{R\,C^{\top} / \bar{R}}$ \Comment{factored, with $\epsilon$-clamps}
\EndIf
\If{$\tau = \textsc{backbone}$}
  \State $M \gets \beta_1 M + (1-\beta_1)\, G$; \quad $U \gets \dfrac{\sqrt{1-\beta_2^{\,t}}}{1-\beta_1^{\,t}}\; M \oslash D$ \Comment{fp32 momentum}
\Else
  \State $U \gets \sqrt{1-\beta_2^{\,t}}\;\, G \oslash D$ \Comment{no momentum buffer}
\EndIf
\State $U \gets U / \max\big(1, \mathrm{RMS}(U)\big)$ \Comment{update clipping}
\State $W \gets \mathrm{bf16}\big(W_{\mathrm{fp32}} - \eta_t\, U + \xi\big)$, \quad $\xi \sim \mathcal{U}\big({\pm}\,\mathrm{ulp}(W)/2\big)$ \Comment{dithered rounding}
\end{algorithmic}
\end{algorithm}

\section{Experimental setup}
\label{sec:setup}

\paragraph{Model.}
We train a decoder-only transformer with two blocks: the first uses a dense SwiGLU feed-forward layer \citep{shazeer2020glu}, the second an MoE layer with 128 SwiGLU experts of hidden width 4096 and top-2 routing. Width is 4096 throughout, with grouped-query attention (32 query heads, 8 KV heads) \citep{ainslie2023gqa}, learned positions, a GPT-2 BPE vocabulary padded to 50{,}304 \citep{radford2019language}, and tied embeddings, giving 6{,}784M parameters in total with about 440M active per token. Two blocks is shallow by intent as well as by budget. It concentrates 95\% of parameters in a single expert bank, the population whose optimizer state we want to stress. It also lets a 6.78B-parameter four-optimizer comparison run on one GPU. The router operates in float32 with input noise $\sigma = 0.5$ at training time and a zero-initialized gate. We use the Switch-style balancing loss with $\alpha = 0.05$ \citep{fedus2022switch} and a z-loss of $10^{-4}$ \citep{zoph2022stmoe}. LayerNorms are kept in float32.

\paragraph{Data and protocol.}
We stream OpenWebText \citep{gokaslan2019openwebtext}, hashing each document into a 95/5 train/validation split so the two sides share no documents. Each run takes 10{,}000 steps at batch size 64 $\times$ 128 tokens, giving 81.9M tokens in a single epoch with no batch repeats. Validation uses the same 64 held-out batches ($\approx$0.5M tokens) for every optimizer. All four optimizers start from one shared initialization and consume identical batches in identical order, under bfloat16 autocast with per-block activation checkpointing \citep{chen2016training} and gradient clipping at 1.0. Learning rates follow a cosine schedule with 3\% warmup. AdamW and SkewAdam use $3 \times 10^{-4}$, Lion uses $1 \times 10^{-4}$ following the $3$--$10\times$ reduction its authors recommend, and Muon uses $0.02$ for matrices with an internal Adam at $10^{-3}$ for embeddings, router and vector parameters. Runs execute on a single NVIDIA H200, whose 141\,GB simply lets us include the AdamW baseline that a 40\,GB device could not hold. Peak memory is read from CUDA's peak-allocation counter. Full hyperparameters are in Appendix~\ref{app:hparams}.

\section{Results}
\label{sec:results}

\begin{table}[t]
\caption{6.78B-parameter MoE, 10{,}000 steps from a shared initialization. State sizes are analytic (Appendix~\ref{app:memory}) while memory and throughput are measured at the final step. Balance loss has a floor of $0.05$ under uniform routing.}
\label{tab:main}
\centering
\small
\begin{tabular}{lrrrrr}
\toprule
Optimizer & State (GB) & Peak mem.\ (GB) & Tokens/s & Val.\ PPL $\downarrow$ & Balance loss \\
\midrule
SkewAdam & \textbf{1.29} & \textbf{31.3} & 5{,}000 & \textbf{108.4} & 0.0505 \\
AdamW    & 50.55 & 81.4 & 4{,}692 & 126.8 & \textbf{0.0502} \\
Muon     & 25.27$^{\dagger}$ & 57.6 & 3{,}409 & 120.2 & 0.0608 \\
Lion     & 25.27 & 56.6 & \textbf{5{,}075} & 393.7 & 0.0537 \\
\bottomrule
\multicolumn{6}{l}{\scriptsize $^{\dagger}$One float32 buffer per parameter. The Adam-style state Muon keeps for embeddings, router and} \\[-2pt]
\multicolumn{6}{l}{\scriptsize \hphantom{$^{\dagger}$}vector parameters adds roughly 0.8\,GB.} \\
\end{tabular}
\end{table}

\begin{figure}[t]
\centering
\begin{subfigure}[b]{0.49\textwidth}
  \includegraphics[width=\textwidth]{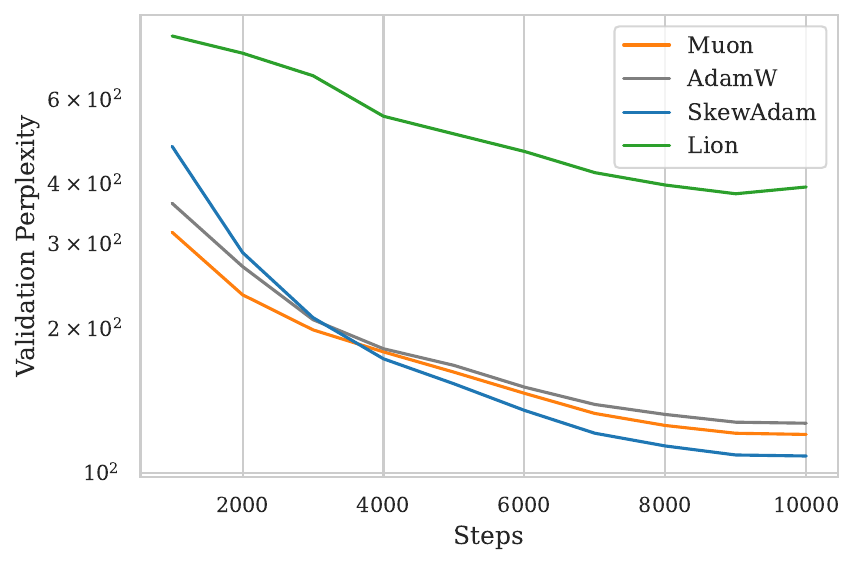}
  \caption{Validation perplexity (log scale).}
\end{subfigure}
\hfill
\begin{subfigure}[b]{0.49\textwidth}
  \includegraphics[width=\textwidth]{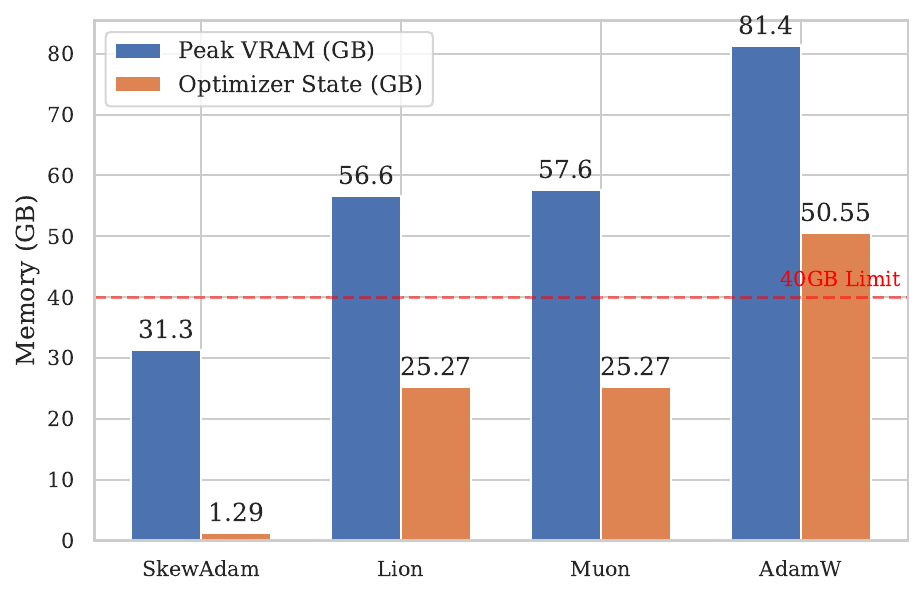}
  \caption{Peak memory and optimizer state.}
\end{subfigure}
\caption{Convergence and memory. AdamW and Muon lead through step 3{,}000 before SkewAdam overtakes both by step 4{,}000 and finishes 14.5\% below AdamW. Only SkewAdam clears the 40\,GB line.}
\label{fig:curves}
\end{figure}

\paragraph{Memory and throughput.}
Table~\ref{tab:main} and Figure~\ref{fig:curves}b give the budget. AdamW peaks at 81.4\,GB, of which 50.6\,GB is optimizer state. The rest is bfloat16 weights (12.6\,GB), gradients (12.6\,GB) and activations. Lion and Muon halve the state and still peak near 57\,GB, above the 40\,GB class of accelerators. SkewAdam's 1.29\,GB of state brings the peak to 31.3\,GB, with headroom to spare on a 40\,GB device. Throughput tracks state traffic. SkewAdam sustains 5{,}000 tokens/s, 6.6\% above AdamW and 1.5\% below Lion, whose single buffer is the cheapest to maintain. Muon pays 32\% relative to SkewAdam for running Newton--Schulz iterations over 6.4B expert parameters every step.

\paragraph{Convergence.}
AdamW and Muon converge faster for the first 3{,}000 steps, trailing AdamW by 114 points of perplexity at step 1{,}000, but SkewAdam passes both by step 4{,}000 and ends at 108.4 against 120.2 for Muon and 126.8 for AdamW. Per-step values are tabulated in Appendix~\ref{app:convergence}. Lion ends at 393.7, more than three times SkewAdam, having peaked at 381.1 at step 9{,}000 and worsened thereafter. Lion does not recover. We read the Lion result as consistent with the magnitude-blindness account, since a fixed-magnitude step treats a heavily routed expert and a rarely routed one identically, while noting it reflects one learning rate and one seed. That SkewAdam ends \emph{below} AdamW is the more surprising outcome. We offer an interpretation rather than a claim. With $\approx$128 tokens reaching each expert per step, Adam's per-coordinate second moments are estimated from little data, while the factored estimator pools statistics over 4096 coordinates per row and column. The pooled preconditioner, plus RMS clipping of occasional large updates, may simply be better conditioned at this routing sparsity.

\begin{figure}[t]
\centering
\includegraphics[width=0.55\textwidth]{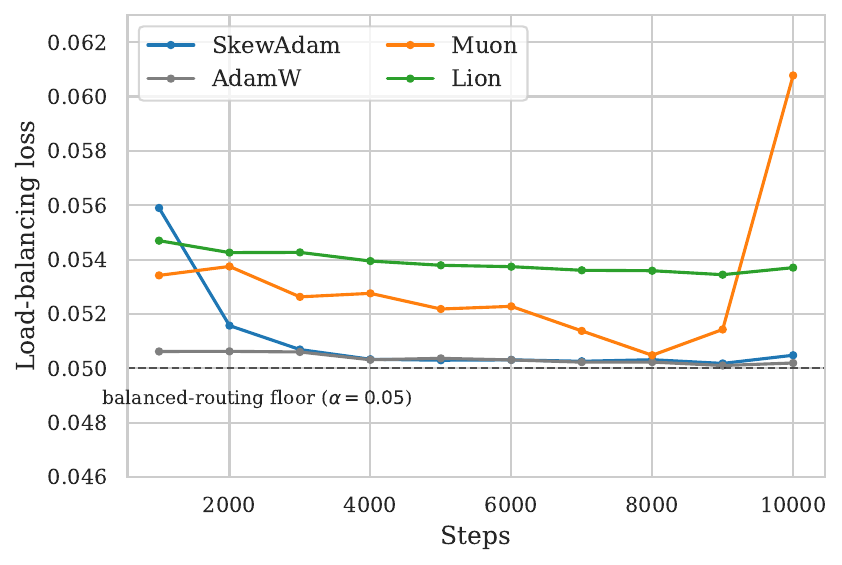}
\caption{Load-balancing loss per evaluation step. The dashed line is the value attained under perfectly uniform routing ($\alpha = 0.05$). SkewAdam and AdamW sit within 1\% of the floor from step 4{,}000, while Muon jumps 22\% above it in the final thousand steps.}
\label{fig:aux}
\end{figure}

\begin{table}[t]
\caption{Factored and low-rank baselines under the identical protocol (same model, data, seed and shared initialization within the batch), run on an NVIDIA H100 NVL 47\,GB MIG slice. Throughput is not comparable with Table~\ref{tab:main} (different hardware). Adafactor's state is analytic. The GaLore-style implementation's state was not instrumented.}
\label{tab:h100}
\centering
\small
\begin{tabular}{lrrrrr}
\toprule
Optimizer & State (GB) & Peak mem.\ (GB) & Tokens/s & Val.\ PPL $\downarrow$ & Balance loss \\
\midrule
SkewAdam & 1.29 & 31.3 & 3{,}778 & \textbf{109.0} & 0.0505 \\
Adafactor & \textbf{0.01} & \textbf{29.6} & 4{,}032 & 149.5 & \textbf{0.0502} \\
GaLore-style (rank 128) & --- & 31.7 & \textbf{4{,}209} & 1{,}839.9 & 0.0510 \\
\bottomrule
\end{tabular}
\end{table}

\paragraph{Adafactor and GaLore.}
The comparison this method most owes its readers is against Adafactor, which supplies the factored estimator SkewAdam builds on. We ran it when compute later allowed, using the same code, data, seed and shared initialization on an H100 MIG slice (Table~\ref{tab:h100}). SkewAdam re-run in the same batch as the anchor lands at 109.0 against its 108.4 on the H200. The two are 0.5\% apart, so the protocol transfers across hardware. Uniform Adafactor carries even less state than SkewAdam, 12\,MB with no momentum anywhere, but plateaus at 149.5, forty perplexity points behind, with its final thousand steps improving by less than 0.1. Both optimizers share the rank-one second-moment estimator and update clipping. The difference is that SkewAdam keeps momentum while Adafactor drops it entirely, alongside Adafactor's annealed decay.\footnote{HuggingFace Adafactor anneals its second-moment decay as $1-t^{-0.8}$ rather than holding $\beta_2 = 0.999$.} The tier ablation below (Table~\ref{tab:ablation}) locates this gap in the momentum and its decay schedule rather than the allocation. Uniform allocation \emph{with} momentum reaches the same perplexity as the tiered policy. The rank-128 GaLore-style baseline in our trainer fails outright at these settings (1{,}839.9), balancing router load like Lion while never learning the language model. We read this as a caution about projecting sparse expert gradients onto low-rank subspaces, not as a verdict on GaLore, whose reference implementation and tuning differ from ours.

\paragraph{Which tier does the work?}
Table~\ref{tab:ablation} toggles the tiers one at a time, on the MI300X under the same protocol. Every variant reaches the same validation perplexity (108.2--108.9, within single-seed noise) and the same load balance of about $0.050$. What moves twentyfold is optimizer state. Restoring momentum to the experts costs 24\,GB and moves perplexity by 0.2, so expert momentum is dead weight and the policy is right to discard it. Factoring the router changes neither perplexity nor balance, so the exact router at 2\,MB is a harmless but not load-bearing choice. Uniform allocation, with momentum and factoring everywhere, matches the tiered policy at twenty times the state. The policy's contribution is therefore memory. It recovers full-momentum perplexity from momentum on the dense backbone alone, where that costs 1.27\,GB rather than 25. These runs also decompose the gaps to the two tuned baselines, all measured on this same MI300X. Uniform allocation \emph{with} momentum reaches 108.3, so the 31-point gap to tuned Adafactor (139.7) is the absence of momentum and its annealed decay rather than the allocation. The 10-point gap to tuned AdamW (118.5) is not momentum either, since AdamW carries momentum on every parameter. What differs there is the factored second moment together with the update clipping that accompanies it, and these runs cannot separate the two. SkewAdam itself reaches 108.9 here, matching its 108.4 on the H200 and 109.0 on the H100. That is a third platform and a second vendor, within noise.

\begin{table}[t]
\caption{Tier ablation on an MI300X (192\,GB), identical protocol, one seed per variant. Each row toggles one decision of the policy. Perplexity and load balance are flat across all four. Optimizer state and peak memory are not.}
\label{tab:ablation}
\centering
\small
\begin{tabular}{lrrrr}
\toprule
Variant & State (GB) & Peak mem.\ (GB) & Val.\ PPL & Balance loss \\
\midrule
SkewAdam (full policy) & \textbf{1.29} & \textbf{31.4} & 108.9 & 0.0505 \\
\quad + momentum on experts & 25.29 & 55.4 & 108.7 & 0.0506 \\
\quad factored router & 1.28 & 31.4 & 108.2 & 0.0505 \\
\quad uniform (momentum + factored) & 25.29 & 55.4 & 108.3 & 0.0503 \\
\bottomrule
\end{tabular}
\end{table}

\paragraph{Tuning the baselines.}
The one quality claim left standing after the ablation was SkewAdam ahead of AdamW and it rested on a single untuned learning rate, so we swept both strong baselines while leaving SkewAdam at its default (Table~\ref{tab:sweep}). Tuning is worth real perplexity to them. AdamW improves from 126.8 at $3 \times 10^{-4}$ to $118.5 \pm 0.5$ at $10^{-4}$ (three seeds), Adafactor from 149.5 to 139.7 at the same rate (two seeds, 0.005 apart). Adafactor's minimum is bracketed on both sides, since $3 \times 10^{-5}$ undertrains to 257.5 and every higher rate is worse, while AdamW's is bracketed only from above. Adafactor's collapse at $3 \times 10^{-5}$ suggests this step budget undertrains at lower rates generally. Neither tuned baseline reaches SkewAdam. Untuned at $3 \times 10^{-4}$, it leads the best AdamW by ten perplexity points, twenty times the seed-level standard deviation, and the best Adafactor by thirty-one. Tuning narrows the headline gaps but does not close them. The persistent AdamW--Adafactor separation (118.5 vs 139.7, both tuned) is the momentum story of Table~\ref{tab:ablation} again, now visible between independent optimizers.

\begin{table}[t]
\caption{Learning-rate sweeps for the two strongest baselines, run on the MI300X under the identical protocol (the Adafactor $3 \times 10^{-4}$ entry is the H100 run of Table~\ref{tab:h100} and the AdamW $3 \times 10^{-4}$ entry reproduces the H200 baseline to within 0.5). SkewAdam is deliberately left untuned.}
\label{tab:sweep}
\centering
\small
\begin{tabular}{lllr}
\toprule
Optimizer & LRs swept & Best LR & Best Val.\ PPL \\
\midrule
AdamW & $10^{-4}$, $3{\times}10^{-4}$, $10^{-3}$ & $10^{-4}$ & $118.5 \pm 0.5$ (3 seeds) \\
Adafactor & $3{\times}10^{-5}$, $10^{-4}$, $3{\times}10^{-4}$, $10^{-3}$, $3{\times}10^{-3}$ & $10^{-4}$ & 139.7 (2 seeds) \\
SkewAdam (untuned) & --- & $3{\times}10^{-4}$ & 108.4--109.0 (3 GPUs) \\
\bottomrule
\end{tabular}
\end{table}

\paragraph{Routing stability.}
The balancing loss $\alpha E \sum_i \bar{p}_i f_i$ equals $\alpha = 0.05$ exactly when both the router probabilities and the realized token assignment are uniform, which makes deviation from $0.05$ a calibrated imbalance measure (Figure~\ref{fig:aux}). SkewAdam and AdamW stay within 1\% of the floor from step 4{,}000 onward (final values 0.0505 and 0.0502). Lion is stable but elevated 7--9\% above the floor throughout, balancing load while failing to learn the language model. Muon improves steadily to 1--3\% above floor and then, in the last thousand steps, jumps to 0.0608, 22\% above. Whether this is the onset of an instability or a transient cannot be settled from one run, but no other optimizer moves by that much at any point in training. The jump coincides with Muon's only flat segment in validation perplexity.

\paragraph{Zero-shot evaluation.}
For completeness we score the final checkpoints with the LM Evaluation Harness \citep{gao2023lmeval} on PIQA, WinoGrande, HellaSwag and ARC-Challenge \citep{bisk2020piqa,sakaguchi2020winogrande,zellers2019hellaswag,clark2018think}. After 82M tokens, three to four orders of magnitude below modern budgets, all four models sit near chance on three of the four tasks and a few points above chance on PIQA (53.5--55.9\%), with no separation between optimizers beyond one to two standard errors (full table in Appendix~\ref{app:downstream}). At this token budget the downstream numbers neither support nor undermine any optimizer. We report them so the perplexity gains are not mistaken for more than they are.

\section{Limitations}
\label{sec:limitations}

The model is two blocks deep. That choice concentrates 95\% of parameters in one expert bank, which is the stress test we wanted, but it leaves untested how tiered allocation behaves when routing decisions compose across many MoE layers. Most numbers are one run per configuration. The sweeps of Table~\ref{tab:sweep} addressed this where it mattered most, tuning both strong baselines over bracketing grids with repeated seeds and replicating SkewAdam's own number on three GPUs, but SkewAdam was not itself tuned, AdamW was not probed below $10^{-4}$ and Lion and Muon keep single untuned rates. The 82M-token horizon and 128-token contexts are small and weight decay was inert in all runs (Section~\ref{sec:method}). Anyone scaling this recipe to production horizons must reintroduce weight decay, fusing the decay term into the float32 update ahead of the stochastically rounded cast so that it survives the bfloat16 ULP, a configuration we have not yet validated at length. The Adafactor and GaLore runs of Table~\ref{tab:h100} were added when compute later became available. They share everything with the main protocol except the GPU. Where configurations repeat across machines they agree closely, with SkewAdam within 0.6 perplexity over three GPUs and AdamW at $3 \times 10^{-4}$ within 0.5 over two. The GaLore number in particular reflects a single untuned configuration of our own implementation and should not be read as more than that. Our AdamW baseline differs from the factored optimizers in two ways at once, the second-moment estimator and the presence of update clipping, so the 10-point gap in Table~\ref{tab:ablation} attributes to the pair rather than to either alone. The tier ablation (Table~\ref{tab:ablation}) likewise rests on one seed per variant and its perplexity spread of 0.6 across four variants is within run-to-run noise. We read it as evidence of memory-at-parity rather than of any perplexity ordering among the variants. It leaves open whether the tiers separate at all under a multi-seed protocol, or at a longer horizon where expert momentum might begin to earn its 24\,GB. Finally, the downstream evaluations are near chance and should be read as a completeness check, not evidence of capability. We plan to validate the approach at larger scales as compute becomes available.

\section{Conclusion}
\label{sec:conclusion}

Spending Adam's full state only on the dense backbone, factored variance on the expert bank and an exact second moment on the router cuts optimizer state by 97.4\% and peak training memory by 61\% on a 6.78B-parameter MoE. A tier ablation shows this costs nothing. The policy reaches the same perplexity and routing balance as a uniform optimizer carrying twenty times the state. The contribution is memory rather than a better optimizer, giving Adam-family quality at 2.6\% of Adam's state, a comparison that survives sweeping the baselines' learning rates. The practical suggestion is to choose the state each parameter group receives from the gradient statistics of that group rather than applying one rule to the whole model.

\section*{Acknowledgments}
This work was self-funded and run on rented GPU time. The compute budget rather than the experimental design set the scale of the study. The author welcomes collaboration or compute support to extend this study to deeper models, longer horizons and multi-seed protocols.

\bibliographystyle{abbrvnat}
\bibliography{references}

\appendix

\section{Hyperparameters}
\label{app:hparams}

\begin{table}[H]
\caption{Complete configuration. All values are as used in the released training script.}
\label{tab:hparams}
\centering
\footnotesize
\setlength{\tabcolsep}{4pt}
\begin{tabular}{llll}
\toprule
\multicolumn{2}{l}{\textbf{Model}} & \multicolumn{2}{l}{\textbf{Training}} \\
\midrule
Width $d_{\mathrm{model}}$ & 4096 & Steps & 10{,}000 \\
Blocks & 2 (dense FFN, MoE) & Tokens & 81.9M (single epoch) \\
Attention heads / KV heads & 32 / 8 & Batch & 64 seq $\times$ 128 tokens \\
Dense FFN hidden (SwiGLU) & 4096 & Microbatch & 8 (8 accumulation steps) \\
Experts / hidden / top-$k$ & 128 / 4096 / 2 & Gradient clip & 1.0 \\
Router noise (train) & $\mathcal{N}(0, 0.5^2)$ & Schedule & cosine, 3\% warmup \\
Balance coef.\ $\alpha$ / z-loss & 0.05 / $10^{-4}$ & Seed & 42 \\
Vocabulary (GPT-2 BPE, padded) & 50{,}304 & Precision & bf16 master weights \\
Positions (learned, max) & 256 & & fp32 norms and router \\
Dropout & 0 & Checkpointing & per block \\
\midrule
\multicolumn{4}{l}{\textbf{Optimizers}} \\
\midrule
AdamW & \multicolumn{3}{l}{$\eta = 3 \times 10^{-4}$, $\beta = (0.9, 0.999)$, $\epsilon = 10^{-8}$} \\
SkewAdam & \multicolumn{3}{l}{$\eta = 3 \times 10^{-4}$, $\beta = (0.9, 0.999)$, $\epsilon = 10^{-8}$} \\
Lion & \multicolumn{3}{l}{$\eta = 1 \times 10^{-4}$, $\beta = (0.9, 0.99)$} \\
Muon & \multicolumn{3}{l}{$\eta = 0.02$, momentum $0.95$, 3 NS steps; internal Adam $\eta = 10^{-3}$} \\
Adafactor & \multicolumn{3}{l}{HF impl.; $\eta = 3 \times 10^{-4}$, no momentum, decay $1{-}t^{-0.8}$} \\
GaLore-style & \multicolumn{3}{l}{rank 128, $\eta = 3 \times 10^{-4}$, $\beta = (0.9, 0.999)$; full Adam off-matrix} \\
Weight decay & \multicolumn{3}{l}{0.05 (0 on router); inert under bf16 rounding, see Section~\ref{sec:method}} \\
\bottomrule
\end{tabular}
\end{table}

\section{Memory accounting}
\label{app:memory}

\begin{table}[H]
\caption{SkewAdam optimizer state by component. The momentum buffer of the dense backbone is, to first order, the entire footprint.}
\label{tab:memory}
\centering
\small
\begin{tabular}{lrlr}
\toprule
Component & Parameters & State kept & Size \\
\midrule
Backbone momentum (fp32) & 341.4M & $m$ & 1{,}302.2 MB \\
Backbone factored 2nd moment & --- & $R, C$ vectors & 0.5 MB \\
Backbone vector params (norms) & 0.04M & full $v$ & 0.2 MB \\
Expert factored 2nd moments & 6{,}442.5M & $R, C$ per matrix & 12.0 MB \\
Router full 2nd moment & 0.52M & $v$ & 2.0 MB \\
\midrule
Total & 6{,}784.3M & & 1.29 GB \\
AdamW on the same model & 6{,}784.3M & $m, v$ & 50.55 GB \\
\bottomrule
\end{tabular}
\end{table}

Parameter counts by tier: token embedding $50{,}304 \times 4096 = 206.0$M (tied with the output head), positions $1.0$M, attention $83.9$M, dense FFN $50.3$M, norms $0.04$M, for a backbone of $341.4$M; experts $128 \times 3 \times 4096^2 = 6{,}442.5$M; router $128 \times 4096 = 0.52$M. Total $6{,}784.3$M, matching the trainer's logged count.

\section{Convergence detail}
\label{app:convergence}

\begin{figure}[H]
\centering
\begin{subfigure}[b]{0.49\textwidth}
  \includegraphics[width=\textwidth]{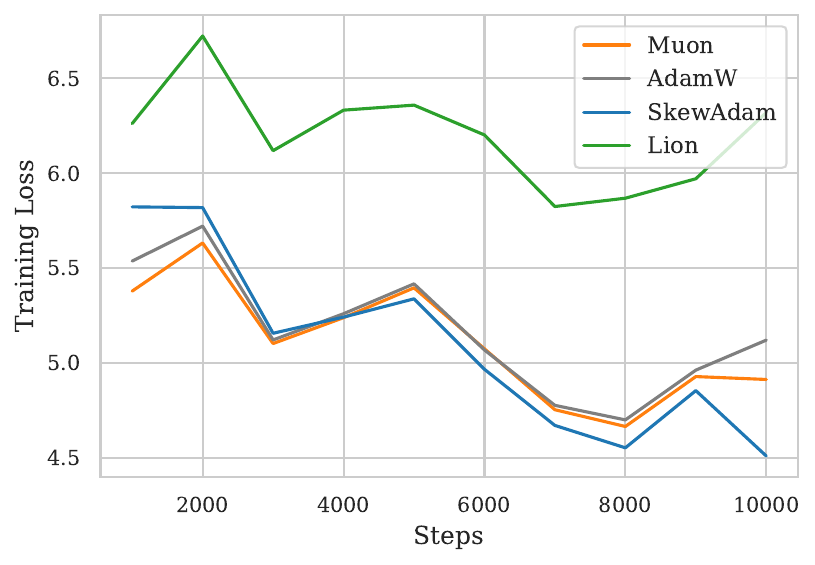}
  \caption{Training language-model loss.}
\end{subfigure}
\hfill
\begin{subfigure}[b]{0.49\textwidth}
  \includegraphics[width=\textwidth]{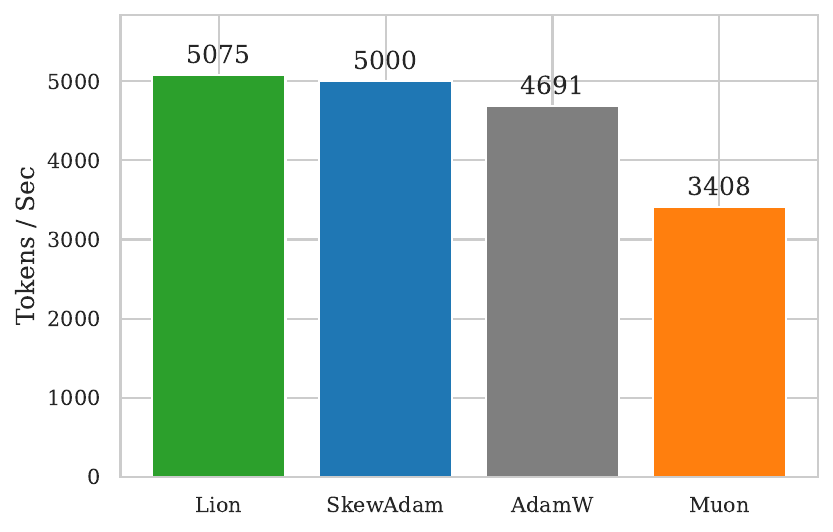}
  \caption{Sustained throughput at step 10{,}000.}
\end{subfigure}
\caption{Training loss and throughput. The loss traces are single-batch measurements at evaluation steps, hence their visible variance.}
\label{fig:appendix-curves}
\end{figure}

Table~\ref{tab:convergence} tabulates validation perplexity at every evaluation step; Figure~\ref{fig:appendix-curves} shows the training loss traces and final sustained throughput. Muon leads for the first three thousand steps, SkewAdam from step four thousand on.

\begin{table}[H]
\caption{Validation perplexity at each evaluation step. Bold marks the best optimizer at each step.}
\label{tab:convergence}
\centering
\small
\begin{tabular}{lrrrrrrrrrr}
\toprule
Step ($\times 10^3$) & 1 & 2 & 3 & 4 & 5 & 6 & 7 & 8 & 9 & 10 \\
\midrule
SkewAdam & 478.4 & 287.6 & 210.4 & \textbf{172.8} & \textbf{153.2} & \textbf{134.9} & \textbf{120.9} & \textbf{113.7} & \textbf{108.9} & \textbf{108.4} \\
AdamW    & 364.1 & 268.9 & 208.4 & 181.3 & 167.4 & 150.8 & 138.8 & 132.3 & 127.5 & 126.8 \\
Muon     & \textbf{316.9} & \textbf{234.8} & \textbf{198.5} & 178.6 & 162.0 & 146.4 & 132.9 & 125.5 & 120.9 & 120.2 \\
Lion     & 812.0 & 747.8 & 671.2 & 552.7 & 507.9 & 466.9 & 421.7 & 397.4 & 381.1 & 393.7 \\
\bottomrule
\end{tabular}
\end{table}

\begin{table}[H]
\caption{Validation perplexity at each evaluation step for the H100 follow-up (Table~\ref{tab:h100}). Bold marks the best optimizer at each step. Adafactor leads for the first two thousand steps and SkewAdam thereafter.}
\label{tab:h100-convergence}
\centering
\footnotesize
\setlength{\tabcolsep}{4pt}
\begin{tabular}{lrrrrrrrrrr}
\toprule
Step ($\times 10^3$) & 1 & 2 & 3 & 4 & 5 & 6 & 7 & 8 & 9 & 10 \\
\midrule
SkewAdam  & 477.3 & 293.8 & \textbf{211.9} & \textbf{174.0} & \textbf{155.1} & \textbf{135.7} & \textbf{121.6} & \textbf{114.4} & \textbf{109.6} & \textbf{109.0} \\
Adafactor & \textbf{403.2} & \textbf{287.0} & 242.7 & 207.3 & 187.1 & 169.2 & 157.8 & 152.4 & 149.6 & 149.5 \\
GaLore & 2703.1 & 2256.8 & 2070.4 & 1965.6 & 1901.7 & 1855.1 & 1829.0 & 1821.7 & 1826.0 & 1839.9 \\
\bottomrule
\end{tabular}
\end{table}

\section{Zero-shot evaluation detail}
\label{app:downstream}

Table~\ref{tab:downstream} expands the summary in Section~\ref{sec:results}. No pairwise difference between optimizers exceeds two standard errors on any task.

\begin{table}[H]
\caption{Zero-shot accuracy (\%) with standard errors, LM Evaluation Harness, batch size 32. PIQA and WinoGrande report accuracy; HellaSwag and ARC-Challenge report length-normalized accuracy. Chance is 50\% for the first two and 25\% for the last two.}
\label{tab:downstream}
\centering
\small
\begin{tabular}{lcccc}
\toprule
Optimizer & PIQA & WinoGrande & HellaSwag & ARC-Challenge \\
\midrule
SkewAdam & 54.7 $\pm$ 1.2 & 49.3 $\pm$ 1.4 & 25.4 $\pm$ 0.4 & 21.6 $\pm$ 1.2 \\
AdamW    & 54.6 $\pm$ 1.2 & 49.4 $\pm$ 1.4 & 25.3 $\pm$ 0.4 & 22.9 $\pm$ 1.2 \\
Muon     & 55.9 $\pm$ 1.2 & 50.4 $\pm$ 1.4 & 25.4 $\pm$ 0.4 & 22.0 $\pm$ 1.2 \\
Lion     & 53.5 $\pm$ 1.2 & 51.1 $\pm$ 1.4 & 25.3 $\pm$ 0.4 & 22.2 $\pm$ 1.2 \\
\bottomrule
\end{tabular}
\end{table}

\section{Reproducibility}
\label{app:repro}

The main comparison ran on a single NVIDIA H200 (141\,GB) using the released single-file trainer and evaluator. Code, per-step training logs and figures are available at \url{https://github.com/nuemaan/skewadam}. The environment is installed with
\begin{verbatim}
pip install torch numpy transformers bitsandbytes datasets lm_eval
\end{verbatim}
and the four reported runs come from one invocation that trains all four optimizers in sequence:
\begin{verbatim}
CUDA_VISIBLE_DEVICES=0 python train.py \
    --optimizers "skewadam,adam,lion,muon"
python evaluate.py runs/best_skewadam.pt   # likewise for the others
python plot_metrics.py
\end{verbatim}
The trainer builds one initialization and one cached batch sequence, reseeds before each optimizer and reuses both for all four, which is what licenses the matched-conditions claim of Section~\ref{sec:setup}.

The follow-up of Table~\ref{tab:h100} ran on a different machine (NVIDIA H100 NVL, 47\,GB MIG slice) under the same protocol:
\begin{verbatim}
python train.py --optimizers "adafactor,skewadam,galore" \
    --dataset-name Skylion007/openwebtext
\end{verbatim}
The dataset flag points at the canonical parquet mirror of the same corpus (newer releases of the \texttt{datasets} library no longer accept the legacy \texttt{openwebtext} id). The document-hash split is unchanged, so the validation set is identical. Metrics and the training log for this run are kept under \texttt{runs/h100/}. The tier ablation and the learning-rate sweeps ran on an MI300X via \texttt{experiments/tier-ablation/run\_ablation.py} and \texttt{experiments/lr-sweep/run\_sweep.py} / \texttt{run\_fix.py}; per-run metrics are under \texttt{runs/amd-ablation/} and \texttt{runs/lr-sweep/}.
The trainer writes per-step metrics, analytic state sizes and peak memory to JSON files under \texttt{runs/}; every number in this paper is read from those files or from the evaluation JSONs. The data split is deterministic (an MD5 hash of each document's first kilobyte selects train or validation), so the validation set is identical across runs and machines.

\end{document}